\documentclass[conference]{IEEEtran}
\IEEEoverridecommandlockouts
\usepackage{orcidlink}
\usepackage{amsmath,amssymb,amsfonts}
\usepackage{algorithmic}
\usepackage{graphicx}
\usepackage{subcaption}
\usepackage{textcomp}
\def\BibTeX{{\rm B\kern-.05em{\sc i\kern-.025em b}\kern-.08em
    T\kern-.1667em\lower.7ex\hbox{E}\kern-.125emX}}
\newcommand{\linebreakand}{%
  \end{@IEEEauthorhalign}
  \hfill\mbox{}\par
  \mbox{}\hfill\begin{@IEEEauthorhalign}
}
\usepackage{amsmath,amssymb,amsfonts}
\usepackage{algorithmic}
\usepackage{graphicx}
\usepackage{textcomp}
\usepackage[linesnumbered,ruled,vlined]{algorithm2e}
\usepackage{balance}
\usepackage{authblk}
\usepackage{booktabs}
\usepackage{orcidlink}
\usepackage[export]{adjustbox}
\usepackage{orcidlink}
\usepackage{amsmath,amssymb,amsfonts}
\usepackage{algorithmic}
\usepackage{graphicx}
\usepackage{subcaption}
\usepackage{textcomp}
\usepackage{balance}
\usepackage{url}

\def\BibTeX{{\rm B\kern-.05em{\sc i\kern-.025em b}\kern-.08em
    T\kern-.1667em\lower.7ex\hbox{E}\kern-.125emX}}

\usepackage[
  backend=biber,
  style=ieee,
  minnames=1,
  maxcitenames=2, maxbibnames=6
]{biblatex}

\begin{document}

\title{SAMURAI: Shape-Aware Multimodal Retrieval for 3D Object Identification}

\author[1,2]{Dinh-Khoi Vo \orcidlink{0000-0001-8831-8846} *\thanks{*Both authors contributed equally to this research.}}
\author[1,2]{Van-Loc Nguyen \orcidlink{0000-0001-9351-3750}*}
\author[1,2]{Minh-Triet Tran \orcidlink{0000-0003-3046-3041}}
\author[1,2]{Trung-Nghia Le \orcidlink{0000-0002-7363-2610}**\thanks{**Corresponding author.}}

\affil[1]{University of Science, VNU-HCM, Ho Chi Minh City, Vietnam}
\affil[2]{Vietnam National University, Ho Chi Minh City, Vietnam}

\affil[ ]{\textit{\{vdkhoi, nvloc\}@selab.hcmus.edu.vn, \{tmtriet, ltnghia\}@fit.hcmus.edu.vn}}

% \author{MAPR Submission ID 28}

\maketitle

\begin{abstract}

% Retrieving 3D objects in complex indoor scenes based solely on a masked 2D image and a natural language description is an extremely challenging task. The ROOMELSA challenge requires retrieval without full 3D context, making it difficult to reason about appearance, geometry, and semantics. This is compounded by distorted viewpoints, textureless masked regions, ambiguous queries, and noisy masks. To tackle these issues, we propose a novel Shape-Aware Multimodal Retrieval for 3D Object Identification (SAMURAI) that combines CLIP-based textual matching, shape-guided re-ranking from masked regions, and a robust majority voting mechanism. To further enhance shape reasoning, we introduce a preprocessing pipeline involving the largest connected component extraction to filter noisy masks and background removal to clean candidate object images. Our method also employs hybrid retrieval strategies to leverage both text and shape information effectively. SAMURAI achieves competitive retrieval performance on the ROOMELSA private test set, highlighting the importance of integrating shape priors with language understanding for accurate open-world 3D object retrieval.
Retrieving 3D objects in complex indoor environments using only a masked 2D image and a natural language description presents significant challenges. The ROOMELSA challenge limits access to full 3D scene context, complicating reasoning about object appearance, geometry, and semantics. These challenges are intensified by distorted viewpoints, textureless masked regions, ambiguous language prompts, and noisy segmentation masks. To address this, we propose SAMURAI: Shape-Aware Multimodal Retrieval for 3D Object Identification. SAMURAI integrates CLIP-based semantic matching with shape-guided re-ranking derived from binary silhouettes of masked regions, alongside a robust majority voting strategy. A dedicated preprocessing pipeline enhances mask quality by extracting the largest connected component and removing background noise. Our hybrid retrieval framework leverages both language and shape cues, achieving competitive performance on the ROOMELSA private test set. These results highlight the importance of combining shape priors with language understanding for robust open-world 3D object retrieval.

\end{abstract}


\begin{figure*}[!t]
    \centering
    \includegraphics[width=\linewidth]{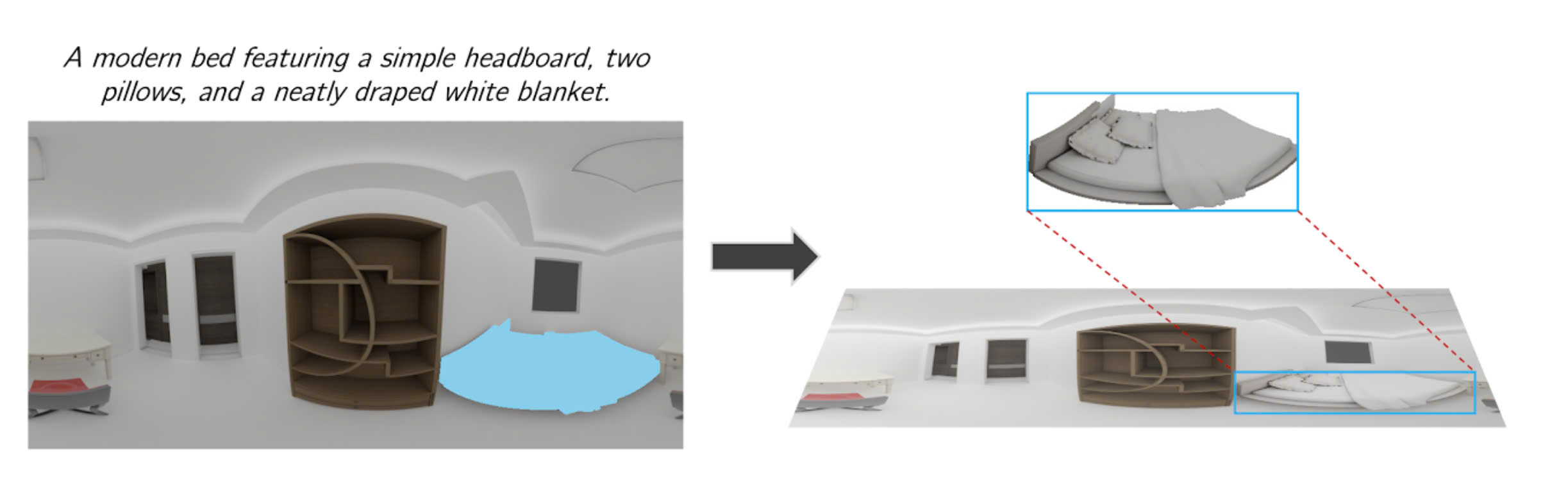}
    \caption{Example input and output of the ROOMELSA challenge.}
    \label{fig:roomelsa}
\end{figure*}
\section{Introduction}

Retrieving 3D objects using natural language and visual context \cite{SketchANIMAR2023, TextANIMAR2023} has become a core challenge at the intersection of computer vision, natural language processing, and 3D scene understanding, with broad applications in robotics \cite{sakaguchi2024simview}, augmented reality \cite{ji2020sketch}, and indoor navigation \cite{dai2017scannet}. Recent breakthroughs in multimodal learning, such as ULIP \cite{xue2023ulip} and OpenScene \cite{peng2023opscene}, demonstrate the potential of aligning 3D, image, and text modalities for open-world 3D recognition and retrieval.

Traditional 3D retrieval systems \cite{houcvpr2020revealnet, chu2022improved} primarily relied on geometric similarity or pre-defined object categories, limiting their flexibility in open-vocabulary settings. Language-grounded 3D understanding methods \cite{chen2020scanrefer, achlioptas2020referit3d}, extend these capabilities by enabling fine-grained object identification through natural language descriptions. However, grounding language in complex 3D environments remains challenging due to intra-class variability, occlusion, and spatial ambiguities.

The ROOMELSA (Retrieval of Optimal Objects for Multimodal Enhanced Language and Spatial Assistance) Grand Challenge \footnote{\url{https://aichallenge.hcmus.edu.vn/shrec-2025/smart3droom}}, held at SHREC 2025, addresses a key problem in 3D scene understanding: retrieving a masked object using only a natural language description and spatial context. Participants must infer and locate the target object without direct visual access, relying solely on linguistic cues and spatial relationships. For example, given a prompt like “a modern bed featuring a simple headboard, two pillows, and a neatly draped white blanket” along with a masked image, the system must identify the corresponding object within a 3D environment. Figure \ref{fig:roomelsa} shows an example.

Recent research highlights the importance of shape priors in resolving ambiguous object references in cluttered scenes. Techniques such as Language-Embedded 3D Gaussians \cite{shi2024language} and Transformer-based grounding methods \cite{chen2022language, NIPS2017Transformer} combine geometric and relational cues to improve retrieval. In the ROOMELSA setting, where only a masked 2D image and a text prompt are available, shape-aware retrieval plays a critical role. This challenge promotes advances in cross-modal retrieval, spatial-language grounding, and comprehensive 3D scene understanding.

To address issues in the ROOMELSA challenge, we introduce the novel SAMURAI framework (Shape-Aware Multimodal Retrieval for 3D Object Identification). Our system leverages the CLIP model \cite{pmlr-v139-radford21a} to align natural language queries with visual features extracted from masked 2D images and candidate object images. To mitigate noise in masked regions, we implement a preprocessing pipeline that extracts the largest connected component and applies padding to ensure complete object capture. We enhance retrieval accuracy by incorporating shape priors through binary silhouette embeddings generated via background removal techniques. The framework employs multiple retrieval strategies, including text-only, shape-only, text-then-shape, and majority voting, to effectively handle diverse query types and scene complexities. By combining semantic matching with shape-guided re-ranking, our system achieves competitive performance on the ROOMELSA private test set, demonstrating the effectiveness of multimodal integration for open-world 3D object retrieval.

Our SAMURAI ranked in the top 5 of the ROOMELSA challenge, achieving an MRR of 0.93 with perfect Recall@5 and Recall@10, comparable to the top team’s MRR of 0.97. Its training-free design ensures robustness across datasets and domains, enabling effective retrieval with language, shape, and visual cues. These results highlight SAMURAI’s competitiveness, offering flexibility and ease of deployment without the need for retraining.

Our contributions are threefold:
\begin{itemize}
    \item We introduce a novel shape-aware retrieval framework that integrates CLIP-based embeddings for natural language and visual features, enhanced by shape priors extracted from binary silhouettes.
    \item We develop a robust preprocessing pipeline designed to handle noisy or fragmented masks, ensuring high-quality cropped object regions for feature extraction.
    \item We present a multi-strategy retrieval approach, combining text-only, shape-only, hybrid text-then-shape, and majority voting methods to address the diverse challenges of the ROOMELSA dataset.
\end{itemize}

\section{Related Work}
\label{sec:related}
\subsection{3D Scene Understanding}

3D scene understanding is critical for applications in robotics and augmented reality. Wang et al.\cite{Wang2024CVPR} proposed unified occupancy for 3D panoptic segmentation, while Delitzas et al.\cite{Delitzas2024CVPR} introduced SceneFun3D with 14.8k interaction annotations for affordance understanding. Wang et al.\cite{Wang2024EmbediedScanCVPR} presented EmbodiedScan, a multimodal suite with RGB-D and language prompts, supporting complex scene interpretation as seen in ROOMELSA. Zhou et al.\cite{Zhou2024CVPR} introduced HUGS, utilizing 3D Gaussian Splatting for holistic urban scene understanding, enabling real-time rendering of new viewpoints and reconstructing dynamic scenes.

Meanwhile, works like ScanRefer~\cite{chen2020scanrefer} and ReferIt3D~\cite{achlioptas2020referit3d} have focused on language-guided 3D localization, with recent improvements such as Chen et al.\cite{chen2022language} enhancing grounding through spatial reasoning. Shi et al.\cite{shi2024language} proposed Language-Embedded 3D Gaussians for open-vocabulary queries, advancing object identification in cluttered environments. 

In contrast, our method introduces a shape-aware retrieval framework that integrates both language and visual features, augmented by shape priors. This approach improves robustness in complex environments like ROOMELSA, where occlusion and ambiguity often pose significant challenges.

\subsection{Multimodal 3D Retrieval}

Recent advancements in 3D scene understanding have significantly contributed to the development of multimodal retrieval systems. ULIP~\cite{xue2023ulip} and MixCon3D~\cite{gao2023mixcon3df} focused on aligning 3D objects, images, and text for open-world retrieval. Yang et al.\cite{Yang2024CVPR} introduced RegionPLC for 3D segmentation, while OpenScene\cite{peng2023opscene} projected CLIP features into point clouds, contributing to cross-modal alignment, which is central to the ROOMELSA challenge. Wang et al.\cite{Wang2024ECCV} explored multimodal Relation Distillation (MRD), a tri-modal pre-training framework designed to transfer knowledge from large Vision-Language Models into 3D representations. Delmas et al.\cite{Delmas2024ECCV}, with their PoseEmbroider approach, combined 3D poses, images, and textual descriptions to create richer human pose representations, surpassing standard multimodal alignment models by effectively handling partial information.

Our proposed method builds on these advancements by introducing a shape-aware retrieval framework that integrates language and visual features, enhanced by shape priors. While existing works like ULIP and MixCon3D focus on aligning 3D, text, and image modalities for general object retrieval, we specifically address the challenges of masked object retrieval in complex environments. We leverage shape priors through binary silhouettes, which is crucial for resolving ambiguities in queries like “the chair near the table,” as highlighted in prior works on fine-grained retrieval~\cite{shi2024language, chen2022language}. By combining multiple retrieval strategies, we offer a more robust solution tailored to the unique demands of the challenge.

\begin{figure*}[!t]
    \centering
    \includegraphics[width=\textwidth]{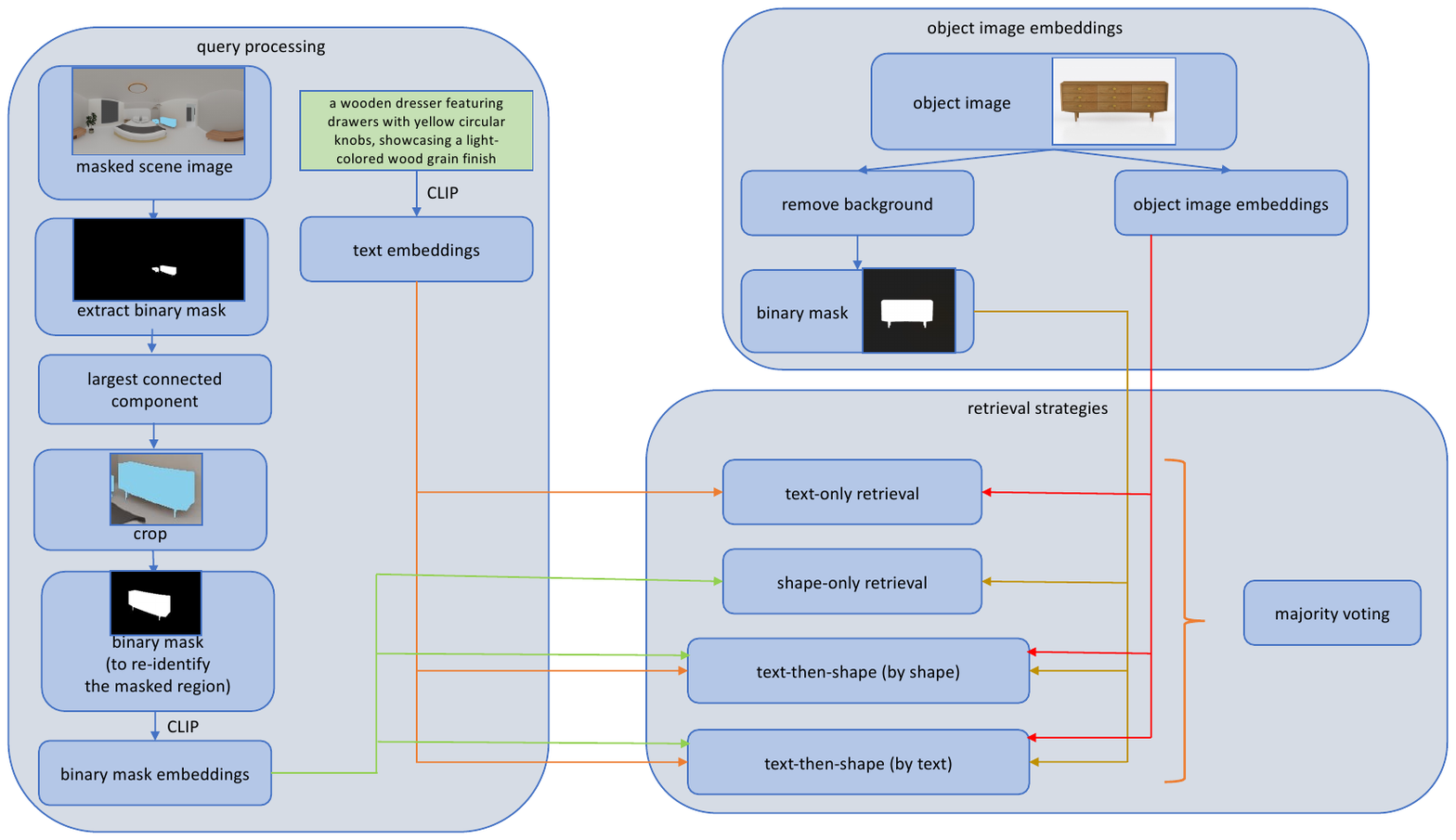}
    \vspace{-20mm}
    \caption{
    Overview of the proposed SAMURAI framework. 
    %\comment{ve xau qua}
    }
    \label{fig:overview}
\end{figure*}
% \raggedbottom

\section{Proposed Method}
\subsection{Overview}

Given a masked 3D scene, a set of candidate objects, and a natural language query, our SAMURAI framework (Figure~\ref{fig:overview}) identifies the target object by jointly reasoning over linguistic descriptions and partial visual cues from the masked region. The system consists of three key components: a query preprocessing module that refines and correctly identifies the mask corresponding to the queried object, a feature extraction stage that encodes language, shape information and object features, and a multimodal retrieval mechanism that integrates textual and visual features to rank the candidate objects.

\subsection{Query Preprocessing}

For each scene’s masked RGB image, our preprocessing module executes four sequential steps. First, it identifies the masked object region using a fixed RGB color (135, 206, 235) and converts it to a binary mask. Second, it extracts the largest connected component via connected component labeling to eliminate noise from multiple masked regions. Third, it applies a 10-pixel padding to the bounding box and crops the region, ensuring the entire object is captured despite discrete masks, such as a bed with segmented parts. Fourth, it re-identifies the masked region in the cropped image using the same RGB color to refine the binary mask, addressing discretization issues. This process generates binary masks, which are subsequently used to extract embeddings and process the query text prompt with the CLIP model.

\subsection{Object Feature Extraction}
% \raggedbottom

We generate two complementary types of embeddings for each object using the CLIP model~\cite{pmlr-v139-radford21a}. First, we produce an RGB embedding that encodes the full object image, capturing appearance, texture, and color details for comparison with the query text prompt. Second, we create a binary silhouette embedding by employing the \texttt{rembg} library to remove the background, isolating the foreground, and converting it to a binary image, focusing on the object’s shape. This background removal eliminates irrelevant contextual elements, while the binary conversion highlights the geometric structure, facilitating robust shape-based comparisons essential for retrieving masked objects in complex 3D scenes.

\begin{figure*}[t!]
    \centering
    \includegraphics[width=\linewidth]{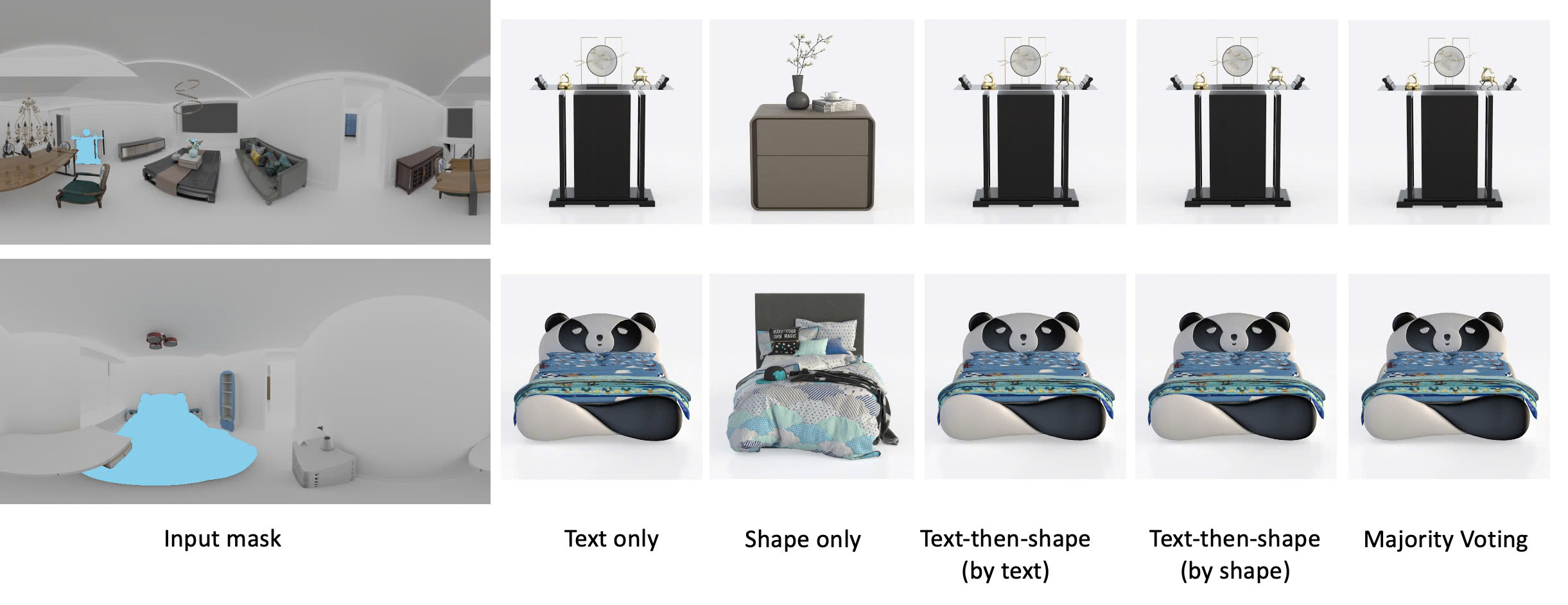}
    \vspace{-6mm}
    \caption{Qualitative results of our five retrieval strategies. 
    %\comment{chu trong hinh qua nho. tang size len cho giong size cua caption} \comment{ca 5 strategies deu ra kq giong nhau nen fig nay khong y nghia gi het}
    }
    \label{fig:qualitative-results}
\end{figure*}
% \raggedbottom

\subsection{Retrieval Strategies}

We propose five complementary retrieval strategies to handle the diverse query types and scene complexities posed by the ROOMELSA challenge:

\begin{itemize}
\item \textit{Text-only Retrieval}: Objects are ranked by computing the cosine similarity between their RGB embeddings and the query text embedding. This strategy emphasizes semantic and appearance cues, making it effective for prompts highlighting color, texture, or material (e.g., “a red wooden chair”).
\item \textit{Shape-only Retrieval}: This strategy matches the binary embedding of the scene’s cropped mask with the binary silhouette embeddings of candidate objects. By focusing on geometric structure, it performs well for shape-driven queries (e.g., “a round table”) where visual cues are limited.

\item \textit{Text-then-Shape (Reordered by Shape)}: Initially selects the top-15 candidates based on text similarity, then refines the top-10 using shape similarity and ranks them by shape relevance. This strategy balances linguistic and geometric information, which is ideal for queries combining appearance and form (e.g., “a tall rectangular cabinet”).

\item \textit{Text-then-Shape (Reordered by Text)}: Also filters candidates via text similarity and refines by shape, but reorders the final list based on text relevance. This variant favors semantic fidelity and suits queries where linguistic precision is more important than geometric fit.

\item \textit{Majority Voting}: Aggregates predictions from all four strategies using weighted voting (e.g., +2 for Text-then-Shape variants, +1 for others). This ensemble method enhances retrieval robustness by integrating complementary strengths across strategies, ensuring stable performance across varied query scenarios.
\end{itemize}

\section{Experiments and Results}
\subsection{Dataset}
The ROOMELSA challenge dataset consists of two main directories: \texttt{scenes} and \texttt{objects}. Each scene folder contains a masked RGB image and an associated natural language query that describes the masked object. In total, the dataset includes 50 distinct scenes.

The \texttt{objects} folder contains candidate 3D objects, each represented by an RGB image, a texture file, and a 3D model in \texttt{.obj} format. For every scene, participants are required to rank the top 10 most likely object matches, from most to least relevant.

\subsection{Metrics}
The evaluation of retrieval performance in the ROOMELSA challenge is based on the following standard metrics:

\begin{itemize}
\item {Recall@k (R@$k$)}: Measures the proportion of queries where the correct object appears in the top $k$ predictions. In this challenge, $k = 1$, $5$, and $10$ are reported.
\item {Mean Reciprocal Rank (MRR)}: Captures the average inverse rank of the correct object across all queries. It is defined as:
$$
\text{MRR} = \frac{1}{\left| Q \right|} \sum_{i=1}^{\left| Q \right|}{\frac{1}{\text{rank}_i}},
$$
where $Q$ is the total number of scenes (queries), and $\text{rank}_i$ denotes the position of the correct object in the ranked list for the $i$-th query. If the correct object is not among the top 10 predictions, the reciprocal rank is set to $0$.
\end{itemize}

\begin{figure*}[t!]
    \centering
    \includegraphics[width=\linewidth]{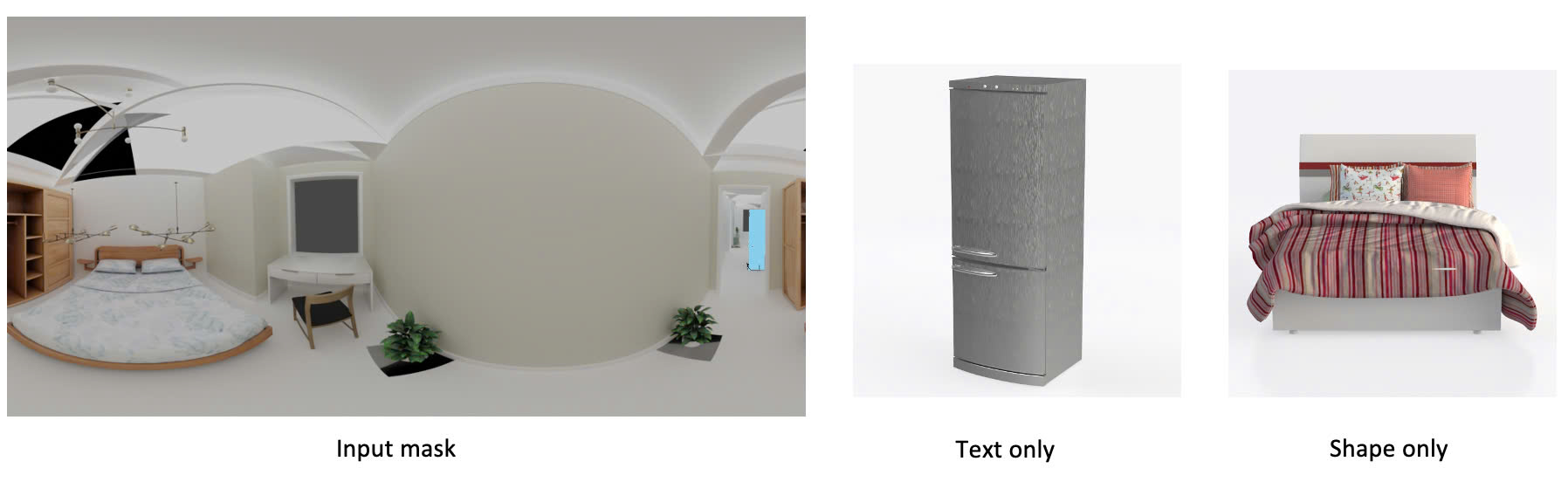}
    \vspace{-7mm}
    \caption{
    Example of the necessity of the combination of text and shape for multimodal retrieval. 
    %\comment{chu trong hinh qua lon. giam size xuong con giong size cua caption}
    }
    \label{fig:text-and-shape}
\end{figure*}

\subsection{Quantitative Results}

Among 18 participating teams in the ROOMELSA challenge, our SAMURAI framework ranked within the top 5. As shown in Table~\ref{tab:roomelsa_results}, our method achieved a high Mean Reciprocal Rank (MRR) of 0.93, with perfect scores in both Recall@5 and Recall@10. Notably, this performance is highly comparable to the top-ranked team's MRR of 0.97, despite differences in methodological complexity.

A key advantage of SAMURAI lies in its training-free design, which eliminates the need for task-specific fine-tuning. This makes our approach robust across different datasets and transferable to new domains without retraining. Its ability to generalize well from language, shape, and visual cues enables effective retrieval even in challenging, unseen scenarios. These results underscore that a lightweight, modular pipeline can remain competitive with state-of-the-art systems while offering greater flexibility and ease of deployment.

\begin{table}[!t]
\centering
\caption{Performance comparison of the top 5 teams in the ROOMELSA challenge using the private test set. Results are evaluated using Recall@1 (R@1), Recall@5 (R@5), Recall@10 (R@10), and Mean Reciprocal Rank (MRR). Higher values indicate better performance. \textbf{Bold} indicates the best score.}
\label{tab:roomelsa_results}
% \resizebox{\textwidth}{!}{
\begin{tabular}{lcccc}
\toprule
\textbf{Team/Method} & \textbf{R@1} $\uparrow$ & \textbf{R@5} $\uparrow$ & \textbf{R@10} $\uparrow$ & \textbf{MRR} $\uparrow$ \\
\midrule
Stubborn\_Strawberries & \textbf{0.94} & \textbf{1.00} & \textbf{1.00} & \textbf{0.97} \\
Ai-Yahh & 0.92 & \textbf{1.00} & \textbf{1.00} & 0.96 \\
MealsRetrieval & 0.92 & \textbf{1.00} & \textbf{1.00} & 0.96 \\
BUCCI\_GANG & 0.90 & \textbf{1.00} & \textbf{1.00} & 0.95 \\
\midrule
 \textbf{SAMURAI (Ours)} & 0.88 & \textbf{1.00} & \textbf{1.00} & 0.93 \\
\bottomrule
\end{tabular}
% }
\end{table}

\subsection{Qualitative Results}

Figure \ref{fig:qualitative-results} presents some of our qualitative results, demonstrating the effectiveness of our retrieval strategies. The queries used for these examples are as follows: "a striking black podium with a sleek glass top, adorned with elegant gold and white accents" and "a cartoon-style illustration of a bed that resembles a panda bear, characterized by its distinctive black-and-white color scheme." The results clearly demonstrate that most of our retrieval strategies, including text-only, shape-only, text-then-shape (ordered by both text and shape), and majority voting, perform well and are competitive in complex 3D environments, successfully identifying the target objects under diverse conditions.

However, there are cases where focusing solely on shape is insufficient, and text descriptions become essential. In complex contexts, the shape alone may not provide enough information to accurately retrieve the object. For instance, Figure \ref{fig:text-and-shape} illustrates a scenario where the query "a sleek, modern refrigerator featuring two doors that open downwards from the top, showcasing a high-end design with a subtle textured pattern on the metal finish" requires reliance on the textual description, as the shape alone may not fully capture the unique features of the object.

\section{Limitations}

Although our method delivers competitive results without requiring retraining, several limitations remain. In scenes where the masked input image contains multiple objects that are spatially close or share similar shapes and textures, the model may struggle to accurately identify and retrieve the intended target. Additionally, highly fragmented or incomplete masks can result in parts of the object being lost during preprocessing. In some extreme cases, the true extent of the object may extend more than 10 pixels beyond the bounding box, negatively affecting retrieval performance. These challenges highlight the need for improved object separation and more robust mask refinement techniques to enhance retrieval accuracy.

\balance
\section{Conclusion and Future Work}

In this paper, we presented a shape-aware multimodal retrieval system that integrates both semantic (textual) and visual (shape) features using CLIP. This design facilitates flexible hybrid strategies and efficient experimentation. While our approach achieves competitive performance without the need for retraining, challenges such as multi-object scenarios and fragmented masks still impact retrieval accuracy.

In the future, we plan to improve retrieval performance by addressing key challenges. For instance, mask fragmentation could be mitigated using advanced object completion or image inpainting techniques. Additionally, integrating deeper spatial context could enhance 3D spatial reasoning, leading to more accurate localization of target objects, especially in complex scenes.
% \raggedbottom

% \comment{neu het cho thi viet tat ten hoi nghi, journal PHO BIEN: CVPR, ...}

% \comment{dung tool de kiem tra viet tieng Anh; prompt: you are US native speaker and expert in computer vison, machine learning. proofread, revise this paper. keep all references, citations ..... dai loai vay}
\section*{Acknowledgment}
This research is supported by research funding from Faculty of Information Technology, University of Science, Vietnam National University - Ho Chi Minh City.

\nocite{*}
\printbibliography

\end{document}